\documentclass[conference]{IEEEtran}

\usepackage{amsmath,amssymb,bm}
\usepackage{siunitx}
\usepackage{graphicx}
\usepackage{booktabs}
\usepackage{placeins}
\usepackage{needspace}
\usepackage{xcolor}
\usepackage[hidelinks,bookmarksopen=false]{hyperref}
\usepackage[capitalize,noabbrev]{cleveref}
\crefname{figure}{Fig.}{Figs.}
\Crefname{figure}{Fig.}{Figs.}
\usepackage[nopatch=footnote]{microtype}
\usepackage{caption}

\newcommand{\norm}[1]{\left\lVert#1\right\rVert}
\newcommand{\simlabel}{\textsc{(sim)}}
\newcommand{\hwlabel}{\textsc{(hw)}}
\newcommand{\columnfig}[1]{\includegraphics[width=0.975\linewidth]{#1}}
\newcommand{\widefig}[1]{\includegraphics[width=0.9804\textwidth]{#1}}

\title{\LARGE \bfseries
Fingers as Legs: Learning Self-Supported Locomotion and Manipulation with an Anthropomorphic Hand
}

\author{
\IEEEauthorblockN{Amirhossein Kazemipour, Hehui Zheng, and Robert Katzschmann%
\thanks{All authors are with the Soft Robotics Lab, ETH Zurich, Zurich,
Switzerland. \texttt{\{akazemi, zhengh, rkk\}@ethz.ch}}}
}

\IEEEoverridecommandlockouts
\IEEEaftertitletext{%
  \begin{minipage}{\textwidth}
    \centering
    \widefig{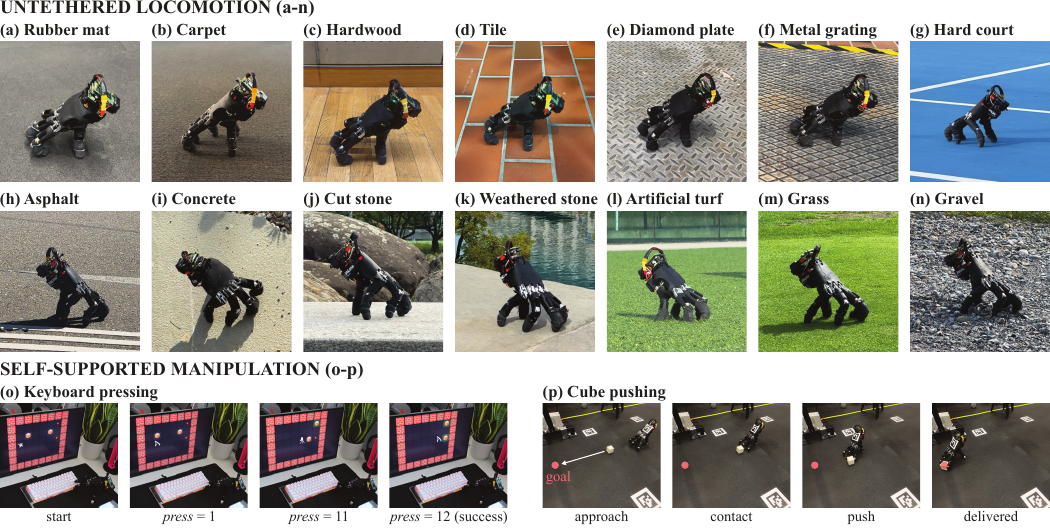}
    \captionsetup{font=footnotesize}
    \captionof{figure}{The hand combines untethered locomotion with
self-supported interaction. Top:
locomotion demonstrated on 14 surfaces: (a) rubber mat, (b) carpet,
(c) hardwood, (d) tile, (e) diamond plate, (f) metal grating, (g) hard court,
(h) asphalt, (i) dry concrete, (j) cut stone, (k) weathered stone, (l) artificial
turf, (m) grass, and (n) gravel. Bottom: self-supported manipulation: (o) 12
keyboard presses completing a Sokoban level; (p) object pushing from approach
through delivery.}
    \label{fig:hero}
  \end{minipage}
  \par\medskip
}

\hypersetup{
  pdftitle={Fingers as Legs: Learning Self-Supported Locomotion and Manipulation with an Anthropomorphic Hand},
  pdfauthor={Amirhossein Kazemipour, Hehui Zheng, Robert Katzschmann},
  pdfsubject={Robotics},
  pdfkeywords={Mobile manipulation, loco-manipulation, legged robots, multifingered hands, reinforcement learning}
}

\begin{document}
\maketitle

\widowpenalty=10000

\begin{abstract}
A walking robotic hand must use the same fingers to move its body, support its weight, and interact with the environment. We show how an anthropomorphic hand can learn these skills while retaining its finger design and position controller. Onboard power and computation make the platform self-contained. Our reinforcement learning approach accounts for the hand's unequal fingers, with training in a simulator calibrated from hardware measurements. In simulation, the hand moves faster with our reward formulation than with tuned rewards originally designed for quadrupeds. On hardware, task-specific policies enable untethered crawling, steering, and fall recovery. While supporting its own weight, the hand also executes successive keyboard commands without vision and pushes an object to targets using overhead visual feedback. These results demonstrate a compact mobile manipulator that reuses its fingers for locomotion and interaction, without a separate locomotion mechanism.
\end{abstract}


\section{Introduction}
\label{sec:intro}

A robotic hand with its own mobility could operate in confined workspaces
without requiring the arm that normally carries it to follow. For example,
a larger robot could place the hand on a support surface near a restricted
opening. The hand could then move toward a control or object, interact with
it, and return for retrieval. Using its fingers as legs, like Thing in
\emph{The Addams Family}, would avoid a separate locomotion mechanism. We study
these capabilities on an off-the-shelf WUJI right hand~\cite{wujiHandDocumentation}
without modifying its fingers or its built-in position controller.

Realizing this idea requires the same fingers to move the body, support its
weight, and interact with the environment. When a finger lifts to step or
press a key, the remaining contacts must keep the hand balanced. Pushing an
object likewise requires maintaining support during task contact. The
opposed thumb and unequal fingers in \cref{fig:hero} complicate this
coordination because each fingertip has a different reach. The palm also
rests at a tilt, so its body frame is not a level reference for commanded
motion.

Our approach has three parts. The locomotion reward is built around the
hand's own stance: a penalty pulls each fingertip toward its nominal stance
position, commands are expressed in a control frame that removes the nominal
palm tilt, and the policy decides where and when each finger steps. Each task
has its own policy, trained in a simulator calibrated to measured fingertip
friction and to the hand's response to filtered position commands. Onboard
power and computation enable untethered operation. We also evaluate
self-supported keyboard pressing without vision and vision-guided object
pushing.

\Needspace{6\baselineskip}
Our contributions are threefold:
\begin{itemize}
  \item \textbf{System:} An \SI{818}{\gram} mobile manipulator built from a
        commercial anthropomorphic hand, retaining its finger kinematics and
        position controller. It carries its own battery and computer, runs
        untethered, and uses the same fingers to walk, to support itself, and
        to interact.
  \item \textbf{Method:} A locomotion reward built around the hand's own
        stance. A footprint objective pulls each of the unequal fingers toward
        its own nominal stance position, while the policy learns where and when
        each finger steps.
  \item \textbf{Evaluation:} In simulation, the hand moves faster with our
        reward than with quadruped rewards adapted to it. On hardware,
        task-specific policies crawl untethered on 14 surfaces, steer, recover
        from falls, press successive keyboard keys without vision, and push an
        object to targets with overhead visual feedback.
\end{itemize}
\section{Related Work}
\label{sec:related}

\paragraph{Mobile hands and finger-limbed robots}
Prior work has explored finger symmetry, modularity, and body reconfiguration
to enable mobility.
DLR-Crawler uses six identical hand-finger modules as hexapod
legs~\cite{gorner2008dlrcrawler}. Gao
\emph{et al.} combine a purpose-built symmetric, reversible hand with
cyclic gaits optimized by a genetic algorithm and vision-guided
manipulation~\cite{gao2026detachable}.
Hand-shaped avatars have been studied for learned locomotion~\cite{sasaki2025hasra}
and detachment from humanoids~\cite{shimobayashi2025handoid}. Other designs
reconfigure hands into humanoids~\cite{li2026handroid} or use origami digits
for grasping and crawling~\cite{lerner2026origami}. Reinforcement learning
also coordinates locomotion and manipulation with reconfigurable
limbs~\cite{sun2023bridging}. We instead keep the unequal finger geometry of
a commercial anthropomorphic hand and add untethered locomotion and
self-supported manipulation.

\paragraph{Manipulation with legs}
A leg used for manipulation is no longer available for support. Robots use legs
to press buttons, open doors, and move
objects~\cite{cheng2023legs,arm2024pedipulate,he2024visual,ji2023dribblebot}.
Whole-body policies coordinate locomotion and manipulation~\cite{fu2022deepwbc}.
Hierarchical navigation systems separate learned skills, perception, and
planning~\cite{hoeller2023parkour}. Our hand must likewise balance interaction
and support, with its fingers providing all body support.

\paragraph{Morphology and reward design}
Locomotion learning often exploits a robot's left--right symmetry through
mirror losses, data augmentation, or equivariant policies~\cite{yu2018symmetric,
abdolhosseini2019symmetric,mittal2024symmetry}. These methods need a mirroring
of states and actions that leaves the reward unchanged~\cite{mittal2024symmetry}.
A hand with an opposed thumb and four unequal fingers has no such mirror.

Gait rewards for legged robots often prescribe timing. Periodic reward
composition assigns each leg swing and stance intervals and relative
phases~\cite{siekmann2021periodic}, and Margolis and
Agrawal~\cite{margolis2023walk} (Walk These Ways) reward a parameterized
contact schedule and Raibert foot-position targets. Our footprint objective
anchors place rather than time: each fingertip is pulled toward its own
nominal position, captured from the settled stance and expressed in a body
frame that is level at that stance, with a weaker pull along the direction of
travel so that steps are cheap. The lift objective only encourages a stepping
rate that grows with the commanded speed; which finger steps when, and how
far, is left to the policy.

\paragraph{Sim-to-real transfer}
Sim-to-real locomotion relies on modeling
actuator dynamics, observation noise, latency, and contact
variation~\cite{tan2018sim,hwangbo2019learning,bjelonic2025systematic}. For our
position-controlled hand, hardware measurements inform the actuation and
contact models.
\section{System and Common Learning Framework}
\label{sec:platform}

\subsection{Self-contained hand platform}
We equip a \SI{738}{\gram} off-the-shelf WUJI right hand for untethered
operation while retaining its finger kinematics, factory controller gains,
and vendor-provided position-control interface. The hand has 20 actuated
joints, four per finger. The joints are non-backdrivable. We configure the
firmware's position-command low-pass filter to a \SI{3}{\hertz} cutoff.
Each joint is limited to
\SI{1.0}{\ampere} during the experiments.

\begin{figure}[!htb]
\centering
\columnfig{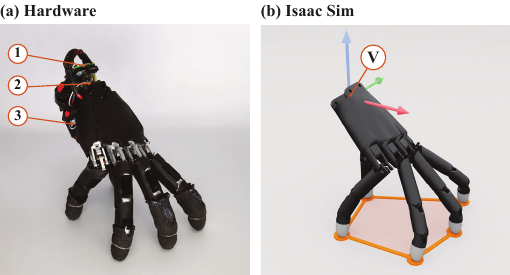}
\caption{Onboard power, sensing, and computation make the commercial hand
self-contained. (a) Hardware with (1) a Raspberry Pi Zero~2 W,
(2) a BNO085 IMU, and (3) a four-cell lithium-polymer battery.
The dorsal module adds \SI{80}{\gram} to the \SI{738}{\gram} hand.
(b) Corresponding NVIDIA Isaac Sim model, showing the stance-calibrated
control frame $V$ (colored axes) and the support polygon formed by the five
fingertip contacts.}
\label{fig:platform}
\end{figure}

The dorsal module in \cref{fig:platform}(a) provides onboard power, sensing,
and computation, bringing the robot's mass to \SI{818}{\gram}.
A ROS~2 stack runs a \SI{500}{\hertz} serial driver and policy inference at
a nominal \SI{50}{\hertz}. The actor and observation normalizer execute on
the Pi as a 32-bit floating-point ONNX model. Crawl commands arrive over a dedicated
\SI{2.4}{\giga\hertz} gamepad link, with Wi-Fi outside the locomotion control
loop. A safety supervisor monitors communication, electrical limits, and
attitude.

\subsection{Policy interfaces}
Each policy is a feedforward network, trained with PPO in simulation
(\cref{sec:locomotion}) and run onboard at \SI{50}{\hertz}. It maps a short
history of proprioceptive inputs, plus the task inputs in
\cref{tab:observations}, to an increment of the joint position targets that
the hand's built-in position controller then tracks.

All policies receive 46 common inputs at each control step: each joint's
angle relative to its fixed reference angle (20, rad), a unit vector
indicating the direction of gravity (3, dimensionless), angular velocity
(3, rad/s), and the previous policy output (20, dimensionless).
The gravity vector and angular velocity are expressed in the palm frame.
The previous output is the clipped policy output $a_t$ defined below.
\Cref{tab:observations} summarizes the additional task-specific inputs.
Each observation term retains eight samples, oldest to newest, before the
term histories are concatenated and normalized using fixed training statistics.
Motor current serves only the safety supervisor.

\begin{table}[t]
\centering
\small
\caption{Task-specific information available to each actor. Dimensions are
in parentheses. $V$ is each task's stance-calibrated frame
(\cref{sec:stance}). Positions are in metres and velocities in m/s.}
\label{tab:observations}
\setlength{\tabcolsep}{3pt}
\begin{tabular}{@{}p{0.18\linewidth}p{0.76\linewidth}@{}}
\toprule
Policy & Task-specific information \\
\midrule
Crawl / recovery & Planar command in $V$ (2), yaw-rate command in rad/s (1) \\
Keyboard & Requested-key indicator (4), pressing fingertip position in $V$ (3) \\
Pushing & Root-relative object position (3), world object velocity (3), object-to-target vector (3), all expressed in $V$ \\
\bottomrule
\end{tabular}
\end{table}

At each control step, the controller adds the joint-target increment
$\Delta q_t^{\mathrm{tar}}=\alpha_q\tanh(a_t)$ to the previous target and
enforces joint limits. Here, $a_t$ is the clipped policy output and
$\alpha_q$ is the task-specific angle scale: 0.060, 0.028, and
\SI{0.040}{\radian} for crawl/recovery, keyboard, and pushing, respectively.
Simulation includes a one-step command delay and a \SI{3}{\hertz} low-pass
filter. Deployment uses the same observation and action transforms, with
additional hardware safety limits.

\subsection{Hardware-calibrated simulation}
Policies act through the hand's retained position controller. We therefore
calibrate the simulator against the measured response to its filtered
position targets.
Frequency sweeps of the proximal and distal interphalangeal
joints, loaded fingertip pulls, and timed command responses identify stiffness,
fingertip friction, delay, filtered joint speed, and filter cutoff (\cref{tab:sysid}).

\begin{table}[t]
\centering
\small
\caption{Hardware measurements used to parameterize the simulator. Joint
stiffness is expressed relative to the nominal, uncalibrated model.}
\label{tab:sysid}
\resizebox{\linewidth}{!}{%
\begin{tabular}{@{}lll@{}}
\toprule
Quantity & Estimate & Procedure \\
\midrule
Joint stiffness scale & 15--22$\times$ (18.7$\pm$2.5$\times$) & joint frequency sweeps \\
Kinetic fingertip friction $\mu_k$ & 0.88 (range 0.80--0.97) & three loaded fingertip pulls \\
Closed-loop delay & $\approx$\SI{19}{\milli\second} & timed command response \\
Filtered joint speed & 2.8 / \SI{2.5}{\radian\per\second} & flexion / abduction ramps \\
Command filter cutoff & \SI{3}{\hertz} & trajectory comparison \\
\bottomrule
\end{tabular}}
\end{table}
\section{Morphology-Adapted Locomotion}
\label{sec:locomotion}

\subsection{Learning problem and optimization}
The crawl policy learns to follow planar-velocity and yaw-rate commands
while supporting the hand on its fingertips. Commands are sampled with
$v_x\in[0,0.16]$ and $v_y\in[-0.16,0.16]$\,\si{\metre\per\second} and
$\omega_z\in[-0.45,0.45]$\,\si{\radian\per\second}. Episodes terminate
upon palm contact, roll beyond $35^\circ$, pitch beyond $30^\circ$, or
timeout. Contact by any other non-fingertip link incurs a penalty.

We train separate actor and critic networks using proximal policy optimization
(PPO)~\cite{schulman2017ppo} in NVIDIA Isaac Lab~\cite{mittal2025isaaclab}.
Both are multilayer perceptrons with exponential linear unit (ELU) activations.
\Cref{tab:training} lists the actor architecture and training settings.
The actor receives the observations
available on hardware (\cref{sec:platform}). During training, the critic also
receives the base state, joint torque, and fingertip contact force. Physics runs
at \SI{200}{\hertz}, with one policy action every four physics steps.
Each PPO update uses 24 steps per environment, five epochs, and four
minibatches, with a clip ratio of 0.2. An adaptive learning rate starts at
$10^{-3}$. We use $\gamma=0.99$, generalized advantage estimation with
$\lambda=0.95$, and an entropy coefficient of 0.005.

Further training adapts the crawl policy to the dorsal module's added load
and contact conditions. Randomization covers fingertip friction, effort scale, and the
payload's horizontal and vertical center of mass offsets (\cref{tab:training}),
as well as palm mass, palm center of mass, and IMU bias. The actuator gains
vary around the calibrated stiffness multipliers (\cref{tab:sysid}).

\begin{table}[!t]
\centering
\small
\caption{Training and objective settings for the deployed crawl policy
\simlabel. Here and in later captions, \textsc{sim} marks simulation results
and \textsc{hw} marks hardware results. The yaw coefficient is introduced by
curriculum. CoM denotes center of mass.}
\label{tab:training}
\resizebox{\linewidth}{!}{%
\begin{tabular}{@{}ll@{}}
\toprule
Setting & Value \\
\midrule
Parallel environments / episode & 4096 / 20\,s \\
Actor hidden layers & 512, 256, 128 (ELU) \\
Actor input / output & 392 / 20 \\
Control rate & 50\,Hz \\
PPO iterations & 8,000 + 5,000 payload/contact adaptation \\
Planar tracking / yaw $\lambda_v/\lambda_\omega$ & $+1.0$ / $0$, then $+0.5$ \\
Objective weights $\lambda_{\mathrm{fp}}/\lambda_{\mathrm{lift}}(k)/\lambda_{\mathrm{dir}}$ & $-30$ / $+6.25\!\rightarrow\!+0.625$ / $-0.5$ \\
Undesired contact & $-1.0$ \\
$v_z$ / $\omega_{xy}$ / action rate & $-2.0$ / $-0.05$ / $-0.01$ \\
Torque / joint acceleration & $-2.5\!\times\!10^{-5}$ / $-2.5\!\times\!10^{-7}$ \\
Tip friction / gain scale & $[0.30,1.30]$ / $[0.78,1.22]$ \\
Payload CoM / effort scale & $\pm$10/15\,mm at 80\,g / $[1.0,1.5]$ \\
\bottomrule
\end{tabular}}
\end{table}

\subsection{Stance-calibrated fingertip objectives}
\label{sec:stance}
\begin{figure}[!t]
\centering
\columnfig{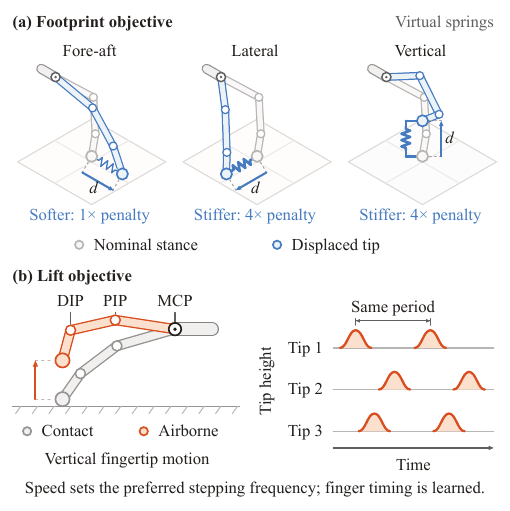}
\caption{The footprint objective penalizes lateral and vertical departures
four times as strongly as fore--aft motion. (a) Virtual springs around nominal
fingertip targets illustrate this ratio for equal displacements. (b) Airborne lifting and illustrative
height traces with the same stepping frequency but different fingertip
timing. MCP, PIP, and DIP denote the metacarpophalangeal,
proximal interphalangeal, and distal interphalangeal joints.}
\label{fig:fingertip_objectives}
\end{figure}

We express fingertip motion in a frame aligned to the nominal stance
and assign each unequal finger its own target.
The footprint objective supplies this geometric reference. Lift and direction
objectives add frequency and swing-direction shaping during training.
The policy learns each finger's timing (\cref{fig:fingertip_objectives}).

\paragraph{Stance-calibrated control frame}
We obtain the reference stance by holding fixed joint targets in simulation
for \SI{4}{\second} with the \SI{80}{\gram} payload, then saving the settled
root pose and joint positions. Let $q_B(t)$ be the root orientation and
$q_{\mathrm{off}}$ the fixed rotation that levels this stance and aligns
forward with the horizontal vector from the palm to the centroid of the
index--little fingers' distal-link origins. We define
$q_V(t)=q_B(t)\otimes q_{\mathrm{off}}$. Frame $V$, illustrated in
\cref{fig:platform}(b), is level and heading-aligned in the nominal stance.
It then follows the root rotation
rather than remaining gravity-leveled. Expressing commands and
base-relative fingertip kinematics in $V$ removes the nominal palm tilt
without suppressing body motion.

\paragraph{Footprint objective}
Each fingertip is referenced to its own nominal position to accommodate the
hand's unequal stance geometry. For fingertip $j$, $p_j$ is its base-relative
position in $V$. Each environment captures $p_j^*$ at its first reward
evaluation after initialization and holds it fixed across subsequent resets.

\begin{samepage}
\noindent With $W=\operatorname{diag}(0.25,1,1)$, the
reward feature is
\begin{equation}
 \bar r_{\mathrm{fp}} = \sum_{j=1}^{5}
 (p_j-p_j^*)^\top W(p_j-p_j^*).
 \label{eq:footprint}
\end{equation}
It contributes $\lambda_{\mathrm{fp}}\bar r_{\mathrm{fp}}$ with
$\lambda_{\mathrm{fp}}=-30$. The penalty has the form of the energy stored in
virtual springs around the stance targets (\cref{fig:fingertip_objectives}(a)):
softer fore--aft springs leave room for stepping, while stiffer lateral and
vertical springs discourage deviations. The stiffnesses are fixed reward weights,
not controller gains; \cref{sec:ablation} sweeps $\lambda_{\mathrm{fp}}$. The targets move with the
body without prescribing ground locations or a footfall sequence.
\end{samepage}

\paragraph{Auxiliary lift objective}
For commanded planar velocity $v_{\mathrm{cmd}}$ and vertical fingertip
velocity $u_j$ relative to the rotating base in $V$, the lift objective favors
a command-dependent stepping frequency (\cref{fig:fingertip_objectives}(b)):
\begin{equation}
 \begin{aligned}
 \phi_t &= \phi_{t-1}+2\pi f(\norm{v_{\mathrm{cmd}}})\Delta t,\\
 z_{j,t} &= (1-\alpha_\tau)z_{j,t-1}
 +\alpha_\tau u_{j,t}e^{-\mathrm{i}\phi_t},\\
 \bar r_{\mathrm{lift},t} &=
 \sum_{j=1}^{5} g_{j,t}z_{\max}
 S(z_{\min},z_{\max},|z_{j,t}|).
 \end{aligned}
 \label{eq:lift}
\end{equation}
The complex exponential moving average uses
$\alpha_\tau=\Delta t/(\Delta t+\tau)$, with control interval
$\Delta t=\SI{20}{\milli\second}$ and $\tau=\SI{0.3}{\second}$.
The phase $\phi$ is kept in $[0,2\pi)$; each wrap marks the start of a new
stepping cycle. Phase, moving average, and contact history reset each episode
and at command resampling.
The actor does not observe $\phi$.
Frequency $f$ linearly maps \SIrange{0.02}{0.16}{\metre\per\second}
to \SIrange{1}{3.5}{\hertz}, with endpoint clamping.
$S(z_{\min},z_{\max},\cdot)$ is a cubic smoothstep that rises from 0 at
$z_{\min}=0.005$ to 1 at $z_{\max}=\SI{0.015}{\metre\per\second}$.

Gate $g_{j,t}$ requires an airborne tip (force below
$F_{\mathrm{th}}=\SI{0.15}{\newton}$), contact at or above $F_{\mathrm{th}}$
since the latest phase wrap, and commanded planar speed at least
$v_{\mathrm{dead}}=\SI{0.02}{\metre\per\second}$.
The lift weight $\lambda_{\mathrm{lift}}(k)$ is 6.25 for the first 18,000
per-environment steps (750 of the 8,000 PPO iterations), then decays
geometrically to 0.625 over the next 36,000 steps and stays there.

\paragraph{Auxiliary direction objective}
For fingertip velocity relative to the rotating base, we penalize its planar
component opposite to the command direction, only when tip force is below
$F_{\mathrm{th}}$ and commanded planar speed is at least $v_{\mathrm{dead}}$.
This leaves planted fingertips free to move backward relative to the base
while propelling it forward.

The full reward formulation also tracks planar velocity and, once a forward
gait has formed, adds yaw-rate tracking at the same 18,000-step point, so that
the policy does not learn stepping and turning at once. It penalizes undesired
contacts,
vertical body motion, roll and pitch rates, joint effort and acceleration,
and changes in action. The planar- and yaw-tracking
kernels use widths of \SI{0.10}{\metre\per\second} and
\SI{0.225}{\radian\per\second}, respectively. \Cref{tab:training} lists all
coefficients.

\subsection{Fall recovery}
Fall recovery lets the hand resume crawling without a person setting it
upright. A separate recovery policy is trained with the same PPO setup; each
\SI{20}{\second} episode starts with the hand lying on its side at a random
heading, and its reward penalizes palm tilt away from level and rewards
reaching the crawl-stance height and joint pose, with no early termination. Once upright, the policy keeps the hand balanced
through continuous joint motion. A smooth joint-target ramp then brings the
hand to rest in the static crawl stance over \SI{1.5}{\second}.
On hardware, an upright-state detector starts this ramp when stance-relative tilt stays below
$10^\circ$ and angular speed below \SI{4}{\radian\per\second} for
\SI{0.5}{\second}. In simulation, the complete procedure rights the hand in 28 of 32 simulated
falls, with a median time of \SI{6.1}{\second} and a mean absolute joint error
of \SI{0.004}{\radian} relative to the crawl stance; in the remaining four
episodes the upright detector did not fire within the \SI{20}{\second}.

\section{Self-Supported Manipulation Policies}
\label{sec:manip}

We test whether the hand can operate a control at a known location and guide
an object using visual feedback. Separate keyboard-pressing and object-pushing
policies retain the calibrated simulator, proprioceptive history, support
requirements, and PPO implementation.
Both use normalized observations and ELU activations, with an actor of three
hidden layers (512, 256, and 128 units) and a critic of three hidden layers
(256, 128, and 64 units) that also receives privileged simulation state during
training.

\subsection{Keyboard pressing without vision}
A key identifier, one-hot during commands and zero while idle, selects
among four learned nominal press locations. Before each evaluation block,
an operator aligns the keyboard to these locations using trial presses.
Encoder-based forward kinematics gives the pressing fingertip position in
palm-attached frame $V$ but does not track hand translation relative to the
keyboard. Successive commands rely on maintained alignment. Misalignment
requires manual realignment.
Rewards encourage approach, increasing press depth, requested-key actuation,
and support from other fingertips. Penalties discourage wrong-key presses,
keycap contact by other hand parts, and loss of stance. Physical keyboard
USB events score presses but are not actor inputs. Latency runs from
command issue to recorded outcome, including event transport and processing.

\subsection{Vision-guided object pushing}
An overhead camera tracks a dorsal marker and the object to construct
the pushing observations in \cref{tab:observations}.
The target remains fixed in the world after each
trial begins. Training uses a
\SI{60}{\hertz} sample-and-hold camera model with a latency of one to three
control steps, 3\% dropout, and \SI{2}{\milli\metre} position noise. The
policy learns approach, contact, and pushing together.

The task rewards approaching the object, moving it toward the target, and
keeping it there, while retaining the support and regularization terms.
\begin{figure}[t]
\centering
\columnfig{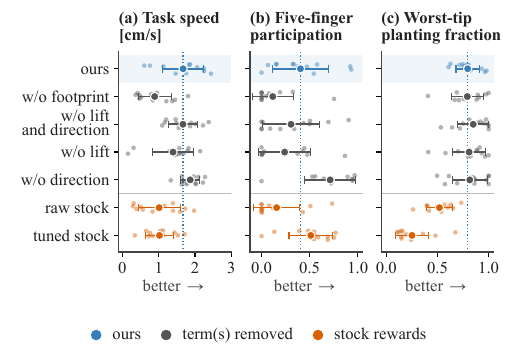}
\caption{Reward-term ablation and stock-reward comparison: the footprint
objective improves task speed and five-finger participation \simlabel. Small
dots show seed means. Large markers and bars
show mean $\pm$ std.\ over twelve training seeds, each evaluated on 256
domain-randomized episodes. Dotted lines mark our formulation.}
\label{fig:ablation}
\end{figure}
\begin{figure}[t]
\centering
\columnfig{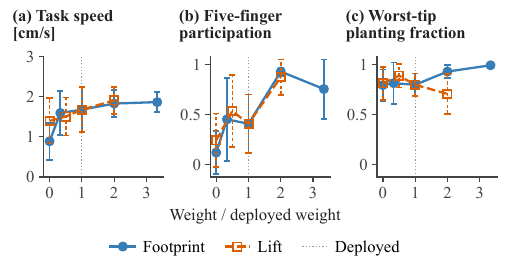}
\caption{Footprint and lift weights change finger participation and contact
posture differently \simlabel. Markers and bars show
mean $\pm$ std.\ across twelve training seeds. Weights are shown as multiples of
their deployed values. The dotted line marks the deployed weight.}
\label{fig:reward_weights}
\end{figure}
\begin{figure*}[!t]
\centering
\widefig{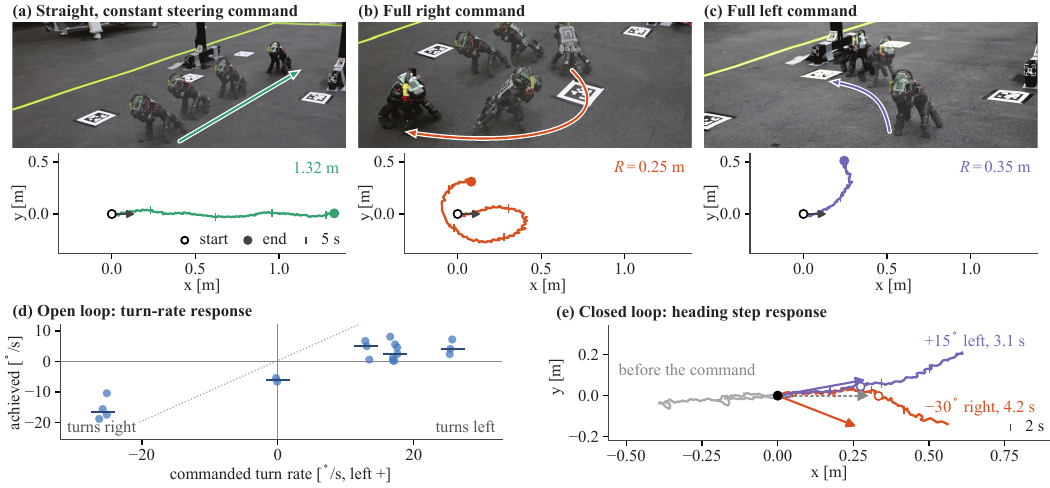}
\caption{The hand turns in both directions with asymmetric rates. IMU
feedback enables commanded heading changes \hwlabel. (a--c) Open-loop snapshots
and paths, aligned to a common origin: (a) corrected straight motion,
(b) the full right-turn command, and (c) the full left-turn command.
(d) Achieved versus commanded turn rate: horizontal bars show medians and the
dotted diagonal marks achieved $=$ commanded. (e) Heading changes measured by
overhead tracking.
The black dot marks the first command. Grey paths precede it. Dotted and solid
arrows show the previous and commanded headings. Circles mark the start of a \SI{1}{\second} interval with the
stride-averaged heading within $\pm 5^\circ$ of the target. Labeled times are
measured from the final command increment.}
\label{fig:steer_hw}
\end{figure*}

\section{Experiments}
\label{sec:results}
\raggedbottom

We first test the reward formulation in simulation, then evaluate how the
untethered hand moves, recovers its stance, and interacts with its surroundings.

\subsection{Reward-term ablation in simulation}
\label{sec:ablation}

We evaluate task speed, five-finger participation, and contact posture.
Participation can include pad-side or nail-side contact, so we interpret it
alongside posture. We compare our formulation with quadruped rewards adapted
to the hand, then remove terms and vary their weights to identify their
contributions.

\paragraph{Design} Only rewards differ across configurations. The hand model,
initial conditions, action interface, observations, curriculum, and domain
randomization are fixed. We train each configuration on the same twelve
seeds with 4096 environments for 8{,}000 crawl iterations. Each policy is then
evaluated in 256 flat-ground episodes with one evaluation seed. Only our
formulation was tested on hardware, after additional adaptation
(\cref{tab:training}).

The seven configurations are our formulation (\cref{sec:locomotion}), four
variants removing the footprint, lift, direction, or both lift and direction objectives,
and two stock baselines. \emph{Raw stock} adapts the rewards from Isaac Lab's
ANYmal-D flat-ground velocity-tracking task~\cite{mittal2025isaaclab}
to the hand, retaining the published weights (feet-air-time $+0.5$,
flat-orientation $-5$).
Both baselines compute the flat-orientation penalty as the sum of squared
horizontal components of unit gravity in the stance-calibrated frame $V$.
It is zero at the nominal crawl stance and independent of heading.
For \emph{tuned stock}, we screen 24 random reward-weight settings on two
seeds for task speed, then evaluate the top three on six fresh seeds.
The selected configuration sets feet-air-time to 2.0 and halves the
flat-orientation weight, rescales the remaining regularization penalties,
and removes the foot-slide and undesired-contact penalties. Six of its twelve reported seeds were also used to select
this configuration. Our weights were fixed by the
hardware campaign before the ablation and were not retuned.

\paragraph{Measures} \emph{Task speed} is progress in the commanded direction,
averaged over each episode and the training command distribution under
domain randomization. \emph{Five-finger participation} is the fraction of
episodes in which every fingertip touches down at least three times and
spends at least 5\% of the episode in contact.

We distinguish tip-side from nail-side contact using the pad axis's tilt
above horizontal toward the nail. Lower tilt indicates more pad-side contact.
We average tilt over time steps in contact and report the worst fingertip and
the mean across fingertips. The \emph{worst-tip planting fraction} is the
fraction of contact time below $64^\circ$ for the fingertip with the largest
fraction of nail-side contact. Contacts are reported per fingertip, not per
side, so tilt serves as the classifier. The \emph{non-nail participation}
variant counts only
contacts below this threshold toward the 5\% contact-time requirement.

\paragraph{Comparison with stock rewards} In \cref{fig:ablation}(a), the hand moves faster with our
formulation than with either tested stock baseline. Compared with tuned
stock, our formulation is 0.65\,cm/s faster on average (95\% confidence
interval [$+0.26$, $+1.02$] across twelve seeds) and faster in 10 of 12 seeds.

Speed alone would hide how the fingers touch the ground
(\cref{fig:ablation}(b) and (c)). Tuned stock reaches a higher mean five-finger
participation, although that difference is uncertain, but it plants fingers on
their nails more in every seed. Our formulation has higher non-nail participation
($+0.35$ [$+0.18$, $+0.53$]), showing why participation and contact posture
must be considered together.

\paragraph{Contribution of individual objectives}
Among the tested reward components, the footprint objective provides the
clearest supported gains in task speed and five-finger participation.
Removing it reduces task speed by
$0.79$\,cm/s [$+0.37$, $+1.19$] and lowers five-finger participation
(\cref{fig:ablation}(a) and (b)). At the deployed weight, this
objective shifts mean contact tilt across fingertips $8.8^\circ$ toward the nail side
([$+1.4$, $+15.9$]). The footprint objective therefore improves
five-finger participation without improving every aspect of contact posture.

The ablations do not establish an independent speed benefit from the
auxiliary lift or direction objectives, whether removed separately or together.
Removing the direction objective alone increases
five-finger participation by 0.31
[$+0.07$, $+0.52$], while its contact-posture effect remains uncertain.
Six additional seed-matched comparisons at doubled footprint weight also leave
the speed and contact effects of direction removal uncertain. These tests do not establish
a benefit from the direction objective. We retain it in the reported reward formulation
because the hardware policies were trained with it.

\begin{figure*}[!t]
\centering
\widefig{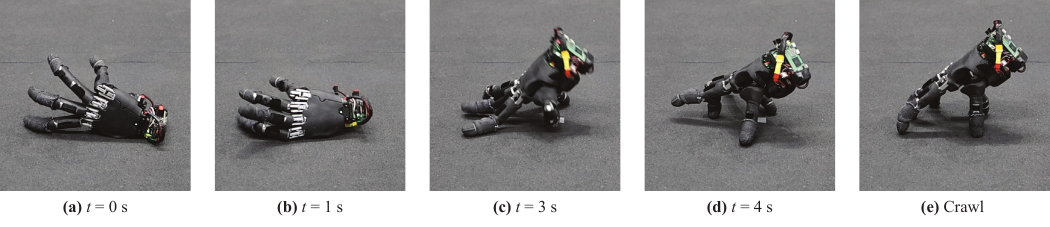}
\caption{The hand returns to its crawl stance after a fall \hwlabel.
(a--d) Learned righting. (e) Controller settling.}
\label{fig:recovery_views}
\end{figure*}

\begin{figure}[t]
\centering
\columnfig{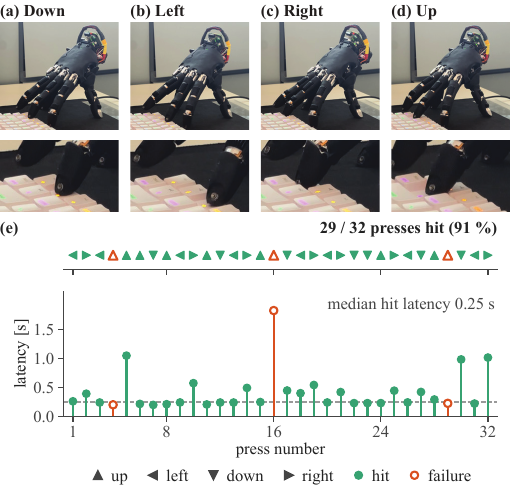}
\caption{The hand presses four arrow keys while supporting itself \hwlabel.
(a--d) Whole-hand views and fingertip close-ups of Down, Left, Right, and Up
presses. (e) Outcomes (top) and command-to-keystroke latency (bottom) in the
32-command sequence. Marker shapes identify commands. Filled green
markers show correct presses (hits), and open orange markers show failures.
The dashed line marks median latency for correct presses.}
\label{fig:keyboard}
\label{fig:keyboard_window}
\end{figure}

\begin{figure}[t]
\centering
\columnfig{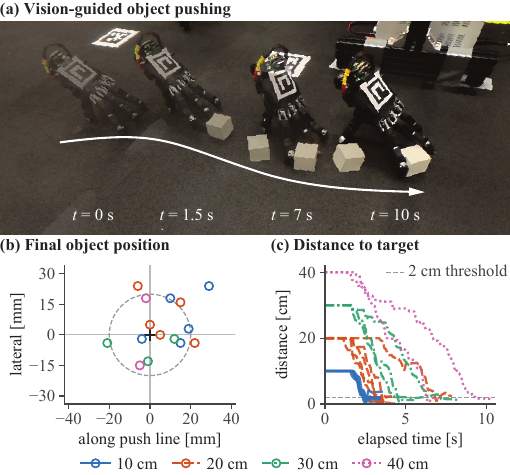}
\caption{One policy approaches and pushes a cube to targets using overhead
visual feedback \hwlabel. (a) A delivery sequence. (b--c) Final target-relative
positions and distance-to-target traces for 15 deliveries, colored by target
distance. Trace time starts at the first recorded control sample; final
positions are the last sample, after the \SI{1}{\second} dwell, during which
the hand keeps pushing. Delivery requires entering the \SI{2}{\centi\metre}
radius (dashed guides), then remaining within \SI{5}{\centi\metre} for
\SI{1}{\second}.}
\label{fig:push}
\end{figure}

\paragraph{Contact posture and reward weights} Doubling footprint or lift
weight raises five-finger participation but affects contact posture differently
(\cref{fig:reward_weights}(b) and (c)).
Doubling footprint weight gives a higher worst-tip
planting fraction than doubling lift weight (paired difference $+0.22$ [$+0.11$, $+0.33$], 10 of 12 seeds). Their speed difference
in \cref{fig:reward_weights}(a) is uncertain
($-0.08$\,cm/s [$-0.39$, $+0.21$], footprint minus lift).

Above the deployed footprint weight, the worst-tip planting fraction keeps
improving while task speed changes remain uncertain; mean five-finger
participation peaks at twice the deployed weight and decreases again by
3.3 times. Higher weights were tested only in simulation. Taken together, the
measured gains in speed and five-finger participation come from the
footprint objective, while the lift and direction objectives shape stepping
rhythm and swing direction during training.

\subsection{Untethered crawling and steering}
The hand crawled untethered under gamepad control across the 14 indoor and
outdoor surfaces in \cref{fig:hero}(a)--(n), from smooth flooring to metal
grating, grass, and gravel; these runs are a qualitative demonstration, shown
in the supplementary video.
We evaluated steering with one policy on a
\SI{1.3}{\metre} mat, using an overhead camera for measurement only.

\paragraph{Open loop} Steering changes the hand's path, but the response
differs between right and left turns. Without a steering command, the hand
drifts right by about $6^\circ$\,s$^{-1}$ regardless of its initial heading.
A constant correction produces nearly straight travel in \cref{fig:steer_hw}(a).
In \cref{fig:steer_hw}(b)--(d), right-turn rate scales with the command, while left-turn rate plateaus.
Across 21 trajectories, mean path speed was
\SI{0.093}{\metre\per\second}.

\paragraph{Closed loop} To close the loop, a proportional controller on the
onboard IMU heading sets the crawl policy's yaw-rate command, adding a constant
offset that cancels the drift and keeping the command within the training
range. \Cref{fig:steer_hw}(e) shows the resulting heading changes. In five
heading-step trials, the hand reached the target heading for both
$\pm 15^\circ$ steps and for the $30^\circ$ right turn. The $30^\circ$ left
turns fell short because the offset that cancels the drift also adds to
left-turn commands, pushing them beyond the limit. These trials demonstrate
commanded heading changes in both directions using onboard sensing.

\subsection{Fall recovery}
The hand recovered in 21 of 25 hardware trials (84\%), including 11 of 14
thumb-side falls and 10 of 11 wrist-side falls. Trials started from fallen
poses with fingers curled and motors initially disabled.
The learned recovery policy raised the hand without assistance; in the four
failures the fingers caught on each other and the hand stalled.
\Cref{fig:recovery_views}(a)--(d) shows a representative sequence.
The upright-state detector then triggered a smooth transition to the crawl
stance shown in \cref{fig:recovery_views}(e), ready for locomotion.

\subsection{Keyboard pressing}
Keyboard pressing tests self-supported interaction with human controls,
as shown in \cref{fig:keyboard}(a)--(d).
With the policy running onboard, an operator issued the four arrow-key
commands in mixed order on a fixed, manually aligned keyboard. We
counted presses of the commanded key as correct and presses of wrong keys
or timeouts as failures.
\Cref{fig:keyboard}(e) shows a sequence containing
29 correct presses from 32 consecutive keyboard commands over \SI{72.5}{\second},
without keyboard realignment.
The hand maintained its support stance with a maximum tilt of
$7.7^\circ$. All three failures were Up-key commands that activated the
adjacent Right Shift key, consistent with a press-location offset.
Median command-to-keystroke latency
among correct presses was \SI{0.25}{\second}.

\paragraph{Keyboard sequences} The same interface controlled Sokoban
through physical keystrokes. With operator-issued
commands, the hand successfully executed the optimal 9- and 12-move solutions for
one- and two-box levels, respectively (\cref{fig:hero}(o)).

\subsection{Vision-guided object pushing}
Object pushing tests whether the hand can coordinate body motion and object
contact while supporting itself.
We used a PLA cube measuring $40\times40\times40$\,mm and weighing \SI{41.4}{\gram}.
With live overhead tracking of the hand and cube, one policy approached and
pushed the cube, as shown in Fig.~\ref{fig:push}(a).
A delivery required entry within \SI{2}{\centi\metre} of the target,
followed by \SI{1}{\second} within \SI{5}{\centi\metre}.
\Cref{fig:push}(b) and (c) show 15 deliveries at target distances of
\SIrange{10}{40}{\centi\metre}, presented as a demonstration of the
capability. Across these deliveries, the final target error averaged
\SI{17}{\milli\metre} (range \SIrange{5}{37}{\milli\metre}), under half
the cube's side length.
\section{Conclusion}
\label{sec:conclusion}

We show that a commercial anthropomorphic hand can serve as a self-contained
mobile manipulator. Its fingers move the body, support its weight, and interact
with the environment, avoiding a separate locomotion mechanism.
A stance-calibrated formulation and a hardware-calibrated
simulator support transfer through the existing position-control interface.
Simulation ablations show that the footprint objective improves task speed
and five-finger participation within the tested formulation.
Untethered locomotion, recovery, keyboard pressing, and vision-guided object
pushing were demonstrated on hardware.

Extending the range of commanded heading changes and integrating onboard
perception would broaden where the hand can operate. Visual registration
could automate keyboard alignment, while onboard hand and object tracking
could support pushing beyond the overhead camera's workspace.

With these improvements, a larger robot could deploy the hand into a confined
workspace, let it operate controls and move objects, and retrieve it. Giving
robotic hands their own mobility could make future robots more versatile in
the spaces and interfaces built for people.

\FloatBarrier
\bibliographystyle{IEEEtran}
\bibliography{refs}

\end{document}